\documentclass[letterpaper, 10 pt, conference]{ieeeconf}

\IEEEoverridecommandlockouts                              
                                                          
\usepackage[T1]{fontenc}
\usepackage{booktabs,multirow}
\usepackage{float,subcaption}
\usepackage{amsmath}

\usepackage[table]{xcolor}
\definecolor{light_gray}{gray}{0.85}

\usepackage{tcolorbox}

\usepackage{siunitx}
\usepackage{etoolbox}
\usepackage{rotating}
\usepackage{hyperref}
\usepackage{cleveref}
\usepackage{makecell}
\newenvironment{packed_enum}{
\begin{enumerate}
  \setlength{\itemsep}{1pt}
  \setlength{\parskip}{0pt}
  \setlength{\parsep}{0pt}
}{\end{enumerate}}

\newcommand{\pseudosection}[1]{\vspace{1.2ex}\noindent\textit{\textbf{#1.~}}}

\title{\LARGE \bf Hazard or Anomaly? Evaluating VLMs for \\Understanding Dangers and Discrepancies}

\author{Murali Indukuri$^{1*}$, Mohammad Eskandari$^{1*}$, Sree Nitya Kollu$^{1}$, Stephanie Lukin$^{2}$, Cynthia Matuszek$^{1\textsuperscript{\textdagger}}$ 
\thanks{* Equal contributions by these authors.}
\thanks{\textsuperscript{\textdagger} This work was supported in part by NSF Grants IIS-2024878 and IIS-2145642, and this material is also based on research that is in part supported by the Army Research Laboratory, Grant No. W911NF2120076.}
\thanks{$^{1}$Interactive Robotics and Language Lab, University of Maryland Baltimore County, Computer Science and Electrical Engineering Department, Baltimore, MD, USA.
    {\tt\footnotesize (muralii1|eskandari|skollu2|cmat)@umbc.edu}}%
    \thanks{\textbf{Accepted to RO-MAN 2026}.}
\thanks{$^{2}$DEVCOM Army Research Laboratory, Adelphi, MD, USA.
    {\tt\footnotesize stephanie.m.lukin.civ@army.mil}}%
    }

\begin{document}
\maketitle
\thispagestyle{empty}
\pagestyle{empty}

\begin{abstract}
Modern safety-critical systems increasingly rely on human--robot interaction to reduce disaster risk and support decision-making during emergencies. Vision--Language Models (VLMs) are promising for these settings because they can interpret complex scenes and communicate safety-relevant information, but they still require careful evaluation to ensure reliable safety reasoning. In particular, current evaluations often frame danger recognition as a binary decision (Safe/Unsafe), making it unclear whether a model is identifying true physical hazards or merely reacting to unusual scene elements. We address this limitation by introducing an explicit distinction between \textit{hazard} and \textit{anomaly}, and by separately recognizing hazardous and anomalous states. We evaluate several state-of-the-art VLMs across two datasets and multiple prompting strategies to test whether this distinction changes model behavior. Our results show that VLMs frequently misinterpret anomalousness as hazardousness, revealing an over-reliance on contextual irregularity as a proxy for danger. We further show that explicitly separating anomaly from hazard provides a more informative evaluation of VLM safety reasoning and exposes failure modes that binary safety judgments can obscure. Our public dataset is available on Roboflow: \href{https://app.roboflow.com/vlm-in-context-anomaly-and-hazard-detection/camera-ready-roman-ds}{https://app.roboflow.com/vlm-in-context-anomaly-and-hazard-detection/camera-ready-roman-ds}.
\end{abstract}

\section{Introduction}





\begin{figure}[t]
    \centering
    \includegraphics[width=0.9\columnwidth]{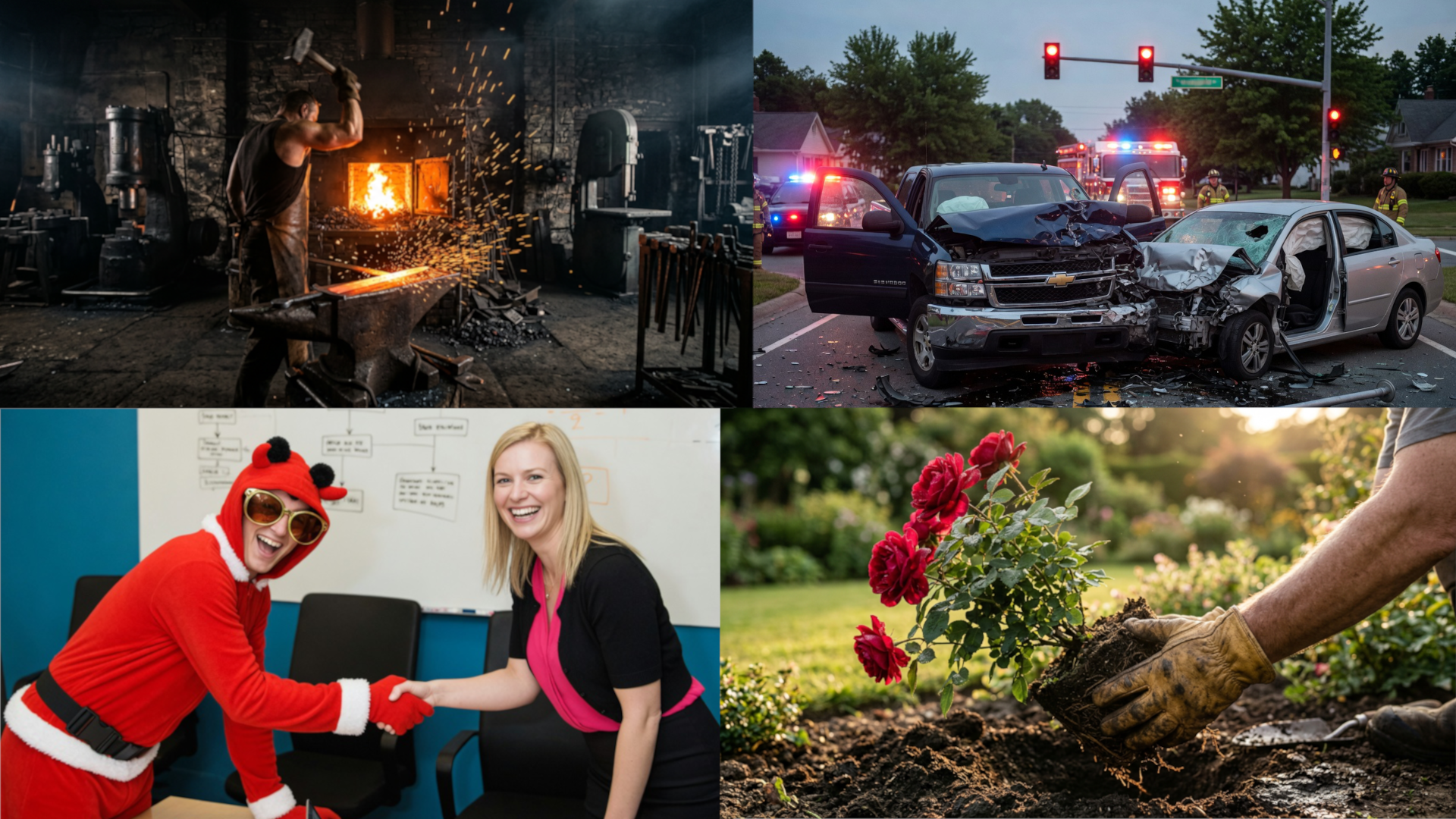}
    \caption{Images extracted from our public dataset representing four classes. Clockwise from bottom right: Safe, Anomalous, Hazardous, and Anomalous-Hazardous. This dataset aims to highlight the conceptual difference  between anomalous and hazardous conditions in borderline cases. \vspace{-1em}}
    \label{fig:safe-unsafe}
\end{figure}

To prevent disasters or respond to them effectively, critical domains such as search and rescue, industrial safety, and space exploration have implemented strict safety standards~\cite{chen_virtual_2018, understandingHRIinIndustry, DIQduringDisaster}. However, ensuring compliance with these standards is still a challenge, in part because operators may still lack adequate situational awareness~\cite{misra2020informationOverload, situatuinal_awareness_2008}. For example, workers in construction zones may overlook abandoned tools, while a first responder scanning an emergency scene may mistake someone's red costume (see ~\cref{fig:safe-unsafe}) for blood, drawing attention away from a real hazard nearby.
As a result, modern safety-critical systems increasingly rely on autonomous robots that can provide situational awareness cues to support human operators in high-risk environments~\cite{industry4, understandingHRIinIndustry}. 
While simple object detection can support Human-Robot Interaction (HRI) in high-risk areas, recent Vision–Language Models (VLMs) appear promising for this role because they can describe and interpret complex scenes by combining visual perception with language-based reasoning~\cite{amaraWhyContextMatters2024, denizRealtimeRoboticsSituation2024, HRIinFutureMilitaryOperations2016, HCIMeetsHRI2015}, limited currently by a lack of clear evaluation of how they form safety judgments and process different types of risk~\cite{bettersafethansorry}.

Although some studies on VLMs report strong performance on multimodal benchmarks~\cite{mmmu}, others document failures even on basic perceptual tasks~\cite{vlmsBlind}; still, whether VLMs can distinguish merely unusual situations from genuine physical danger, or understand the context when making safety judgments, remains an open question. In many evaluations, danger detection is framed as a binary choice~\cite{yang2025coinco, bettersafethansorry}, forcing ambiguous scenes into coarse categories. As a result, anomalous but non-dangerous situations are often labeled as hazardous, while subtle hazards are overlooked. We argue that this stems from the absence of an explicit separation between anomalies and hazards. Following OSHA~\cite{osha2018hazard}, we define a hazard as a source of potential physical harm under normal interaction conditions. An \textit{anomaly}, in contrast, is a violation of expected spatial, functional, or contextual relations that does not necessarily imply danger. These definitions allow us to test whether models can differentiate the orthogonal classes.

We contribute the following:
\begin{packed_enum}
    \item An evaluation pipeline for hazard/anomaly classification in VLMs, designed to study distinctions relevant to situational awareness in safety-critical settings.
    
    \item A curated 610-image benchmark with a four-class hazard–anomaly scheme for evaluating anomaly–hazard separation in Vision–Language Models.
    
    \item A systematic evaluation showing that state-of-the-art VLMs conflate anomalousness with hazardousness under binary safety prompting, and that explicit separation improves calibration.
\end{packed_enum}

    
\section{Related Work}

The concepts of hazard and anomaly are semantically close. In other words, it is essential for the VLMs to disambiguate between these concepts to effectively execute the task. Previous research has shown that it is possible to reliably determine the meaning of words from context~\cite{yaeLLMWordSenseDisambiguous2025, sumanathilakaGPTWordSenseDisambiguous024}, and it is possible for multimodal models to correctly map ambiguous words to the corresponding visual stimuli~\cite{kritharoulaMLLMWordSenseDisambigous2023}, showing that VLMs combined with word sense disambiguation are promising in distinguishing these concepts.
Another relevant area is the study of object-contextual reasoning in multimodal models. Previous research has evaluated the perception and reasoning capabilities of VLMs using datasets such as PAM and CVR~\cite{COCO-OOC-Dataset, wengCaptionThisReason2025}, or has focused on the evaluation of the sensitivity of VLMs to the composition of the captions, including the presence of words that are placed in the wrong position~\cite{winoground_2022}. The results of these studies demonstrate that contextual information and fine-tuning strategies improve contextual reasoning~\cite{VLMAnomalyDetection2025, xuVLMTaskagnosticVideoLanguage2021}.

Existing research has shown that modality bias exists in multimodal reasoning. For example, if textual tokens carry answers to a particular query, vision-language models (VLMs) often rely on textual information over visual information~\cite{amaraWhyContextMatters2024}. In addition, for tasks requiring precise joint reasoning over both textual and visual information, VLMs significantly underperform human baselines~\cite{wadhawanConTextual2024}. Our research attempts to improve the existing models by providing definitions for ``anomaly'' and ``hazard,'' thus encouraging models to focus on relevant visual information while avoiding ambiguities. We have also designed a task where dense captions are used as an input modality to isolate reasoning ability with text-only input.

Other work has explored multimodal reasoning models for robotic applications. For instance, MLLMs (Multimodal Large Language Models) have been incorporated into robots to identify semantic anomalies arising from out-of-distribution interactions~\cite{elhafsiSemanticAnomalyDetection2023}. Research on understanding road accidents has shown that VLMs are well suited for visually grounded perception tasks, while accident prevention is still beyond their capabilities~\cite{kim2025vruaccidentVLMBench}. Other studies have shown that using smaller models can improve their performance over larger models, especially in hazard detection~\cite{xiaoHazardVLMVideoLanguage2025}.

Most relevant to the current research, existing research has explored VLMs for reasoning about anomalies, context, and other related issues. For example, models can generate descriptions for anomalies from images, which can then be used to classify the images with high accuracy~\cite{ZhuLLMUnderstandVisualAnoms}. In addition, BERT models can be used to establish whether an object is within a particular scene, thus providing information for context detection~\cite{yang2025coinco}. 

Misaligned reasoning is an enduring concern in the deployment of such systems. Empirical studies have demonstrated that multimodal models are prone to producing harmful and misleading advice in the process of joint reasoning over images and text data. Open-weight models are particularly susceptible to such model failures~\cite{rottgerMSTSMultimodalSafety2025}. In other studies, the over-reaction problem is reported, where the model is prone to over-reacting and incorrectly identifying scenes that are merely anomalous as dangerous in the process of binary safety labeling,~\cite{bettersafethansorry} in the study of VLMs. 

This study extends the aforementioned research by clearly defining the concepts of anomaly and hazard as distinct labels, allowing the model to reason over contextual anomalies. By including the formal definitions of the concepts within the prompts, the study attempts to address the overreaction problem and provide more consistent results in the identification of the most relevant safety risks.

\section{Methodology}

For a systematic evaluation of current VLM capability in distinguishing the semantically close concepts of hazard and anomaly, we construct a 610-image dataset labeled with one of four classes: Safe, Anomaly, Hazard, or Both. We then use this dataset with several different prompting techniques to evaluate VLM capability. Further, we introduce a simplified classification task that evaluates whether current models can perform the four-class classification given the dense captions associated with a particular image of the dataset.
We begin by formalizing this methodology.

\pseudosection{System Design} For our formalization, we use the following symbolism:
\begin{itemize}
    \item Let $\mathcal{P}$ denote the prompt, an ordered tuple which may be composed of the following elements:
    \begin{quote}

    \begin{itemize}
    \item Let $\mathcal{Q}^T_i$ and $\mathcal{Q}^I_i$ be the $i_{\text{th}}$ set of text and image tokens in $\mathcal{P}$ that the VLM uses as input.
    \item Let $\mathcal{R}_i$ be the $i_\text{th}$ element of $\mathcal{P}$ that represents a VLM response.
    \end{itemize}
    \end{quote}
\end{itemize}
A given element of $\mathcal{P}$ may be associated with a classification code $c_i$, discussed further in the following sections. 

We treat a given VLM agent as a function $\mathcal{A}$ that maps $\mathcal{P}$, a prompt of length $i$, to some response $\mathcal{R}_{i+1}$ (i.e. $\mathcal{A}(\mathcal{P})=\mathcal{R}_{i+1}$). $\mathcal{R}_{i+1}$ may be appended to $\mathcal{P}$ along with other elements, $\mathcal{Q}^T_{i+1}$ and $\mathcal{Q}^I_{i+1}$. Generally speaking, the objective of the VLM is to produce the correct $c_{i+1}$ in $R_{i+1}$ to match the expected in $c_i$ in $\mathcal{Q}^T_i$ or $\mathcal{Q}^I_i$ (i.e. the last element of the prompt $\mathcal{P}$).

\pseudosection{Dataset Design} \label{sec:dataset} Our goal is to evaluate the ability of state-of-the-art VLMs in distinguishing hazards, anomalies, and benign scenes. To this end, we developed a dataset where each image, $\mathcal{Q}^I_i$, is labeled with $c_i$ taking two-bit values corresponding to the following:

\begin{itemize}
    \item \textbf{Safe}: No immediate hazard is present and object-context relations are not violated.
    \item \textbf{Anomalous}: A scene element appears out of place or inconsistent with expected relations (interposition, support, probability, position, or familiar size), but poses no immediate physical risk.
    \item \textbf{Hazardous}: A visible source of immediate harm is present, consistent with OSHA's definition of a hazard, without any contextual irregularity.
    \item \textbf{Anomalous Hazardous}: The scene presents both an immediate hazard and an object--context violation.
\end{itemize}

Unlike traditional safety datasets, which combine contextual deviation and physical danger under a unified label, our two-bit encoding treats the notions of anomaly and danger as orthogonal dimensions. This allows for a controlled assessment of whether predictions are driven by object-level risk cues or more general contextual deviations. This dataset, therefore, allows for a more in-depth study of VLMs' reasoning abilities when judging hazards.

The main dataset comprises artificially generated images based on textual descriptions from several public datasets and images. These images spanned many fields including construction sites, natural disasters, and hazardous objects like weapons~\cite{ds_aerial,ds_ai_safety,ds_chem_spill,ds_firefighter,ds_interior_cv,ds_natural_disaster,ds_ppw_cv}. We believe that the range of environments and contexts that these images were pulled from supports a rigorous evaluation of how VLMs may perform when deployed in real-life anomaly and hazard detection scenarios.

To construct the datasets, images were hosted on Roboflow~\cite{roboflow}. The source datasets were imported to a combined dataset, which we used to construct text descriptions. Images for the final published dataset were generated from these descriptions. In the final dataset, the researchers independently labeled each image with a binary hazard and anomaly labels in the form of \texttt{AH}. For instance, \texttt{01} indicated the image contained hazards but not an anomaly. The researchers reviewed these independent annotations collectively to reach a unanimous agreement.

In borderline cases, a displaced element could be considered either safe or risky, which can cause disagreements among the annotators. Such discussions highlighted the ambiguity between anomalous and hazardous conditions, resulting in more refined definitions of anomaly and hazard. When an agreement could not be reached after a discussion among the annotators, the images were excluded.

The final dataset included over 100 images per class with 610 images in total. The safe, anomalous, hazardous, and anomalous-hazardous categories had 213, 113, 111, and 173 images respectively. This imbalance results from the less common nature of exclusively anomalous or hazardous situations. The study down-sampled to use an equal number of images in all classes to limit testing bias.


\pseudosection{Dense Captioning} We hypothesize that the main bottleneck for VLM performance in our task is the need to appropriately process the context in an image to get the information needed for classification. Work has suggested that current models struggle with this~\cite{ChenSpatialReasoningHard}. We therefore propose the related task of classifying dense captions of images from our dataset into the four discussed classes. With this, we seek to isolate the task of identifying context in an image from reasoning about said context. Based on preceding work that shows VLMs' capability for 3D reasoning from text~\cite{ChandhokSceneGPT}, we hope this results in more accurate classifications for our task.

To generate dense captions, $\mathcal{Q}^T_i$, for each image of our dataset, we use Google's Gemini 3 Flash. Each dense caption is roughly a paragraph long and stored with the associated image's classification code $c_i$. Because our goal is to isolate VLMs' ability to process context, we reviewed the dense captions to ensure they were accurate for the dataset images and that they are sufficiently detailed to be able to justify a given class prediction $c_{i+1}$ for caption $\mathcal{Q}^T_i$.


\pseudosection{Task Design}

\newcommand{\qdef}{\mathcal{Q}^T_0}

\newcommand{\qtf}{\mathcal{Q}^T_f}

\begin{tcolorbox}[title=Core prompt and problem definition $(\qdef)$:]
You are inspecting an environment for anomalies and hazards to help first responders. Use the following definitions. 
\begin{itemize}
    \item Hazard: A scene is hazardous if normal interaction with at least one of the present elements would realistically cause immediate physical harm to the people or put the existing property in significant danger.
    \item Anomaly: A scene is anomalous if any visible element clearly violates normal expectations for this environment (location, arrangement, state, or use).
    \item Safe: Safe scenes include normal, expected human activities in the specifically given appropriate environment, provided no immediate physical hazard is present.
\end{itemize}
\end{tcolorbox}

\begin{tcolorbox}[title=The output structure specification ($\qtf$)]
Decide hazard and anomaly labels independently. Obey this STRICT output format: `\textless a brief, explicit justification for the final decision\textgreater: \textless code\textgreater' Where \textless code\textgreater is ONLY one of: 00 = Safe, 10 = Anomalous, 01 = Hazardous 11 = Anomalous AND Hazardous - No extra paragraphs, no bullet points, no additional commentary.
\end{tcolorbox}

The primary objectives of our task design are to evaluate the ability of modern VLMs to recognize anomalies and hazards, and to determine how well they distinguish between these two closely related concepts. To this end, we define several related tasks in which a VLM agent maps a prompt $\mathcal{P}_i$ to a response $R_i$ containing a classification code $c_i$. For each task, the predicted code $c_i$ is compared against the ground-truth label associated with the target query instance. This general formulation allows us to keep the query content fixed across tasks while varying the prompting strategy, thereby isolating the effects of prompt design and input modality from those of contextual variation between prompts.

For all tasks, we begin by defining the role of the agent and the definitions it must use in the first element of the prompt, $\mathcal{P}$. We denote this $\mathcal{Q}^T_0$.

We include formatting information near the end of each prompt to standardize model responses. We label this $\mathcal{Q}^T_f$, which contains the specification for the code, and how the model must justify its response.
The specific structure for the rest of prompt depended on the prompting technique and input modality as is discussed in the following sections.


\pseudosection{Approach: Zero Shot Prompting} We first test the simplest approach to our classification task: zero shot prompting. This technique allowed the most flexibility for model reasoning and interpretation of the definitions in $\qdef$. Our prompt was of the following form: $\mathcal{P}=(\mathcal{Q}^T_0, \mathcal{Q}^T_1, \mathcal{Q}^T_2, \mathcal{Q}^T_3, \mathcal{Q}_4)$, where the queries 1-3 were:
\begin{tcolorbox}

\begin{itemize}
    \itemsep5pt
    \item[$\mathcal{Q}^T_1$:] Identify concrete sources of immediate harm.
    \item[$\mathcal{Q}^T_2$:] Independently assess contextual irregularities.
    \item[$\mathbf{\mathcal{Q}^T_3}$] $=\qtf$ 
\end{itemize}
\end{tcolorbox}

As mentioned, our study included two variations of the experiment for each prompt: using images from our dataset, or dense captions for those images. $\mathcal{Q}_4$ represents the image for the former experiment and the set of dense captions for the latter experiment.



\pseudosection{Approach: Few Shot Prompting} Because interpreting the definitions for an anomaly vs. a hazard can be subjective, we propose that more sophisticated techniques will improve performance. Few-shot prompting builds on the zero shot prompt by providing example responses rather than the broad instructions in the previous technique. We use the following structure: 
\begin{align*}
    \mathcal{P} = (& \qdef + \qtf, \mathcal{Q}_1, \mathcal{R}_1, \\
                   & \qdef + \qtf, \mathcal{Q}_2, \mathcal{R}_2, \\
                   & \qdef + \qtf, \mathcal{Q}_3, \mathcal{R}_3, \\
                   & \qdef + \qtf, \mathcal{Q}_4, \mathcal{R}_4, \\
                   & \qdef+\qtf, \mathcal{Q}_5).
\end{align*}

Queries $\mathcal{Q}_1$ to $\mathcal{Q}_4$ represent example images or the corresponding dense captions from each class of our dataset, and $\mathcal{R}_1$ to $\mathcal{R}_4$ represent examples of responses the agent should provide for the respective query. $\mathcal{Q}_5$ is the actual image or caption the model must respond to.

The four examples did not have any common elements; they were from unique scenes, domains, and had unique ground truth labels. This keeps the input token usage fairly low to meet restrictions of smaller models while also providing a broad range of examples that should help improve classification performance.

\pseudosection{Approach: Chain-of-Thought Prompting} In addition to providing selected examples, it is possible that guiding agents to identify relevant context from images before classification can improve performance. To facilitate completing our task in a step-by-step manner, we tested each agent with chain of thought prompting. We use the following inductive implementation:
\begin{align*}
    \mathcal{R}_{i+1} &= A(\mathcal{P}_i) \\
    \mathcal{P}_{i+1} &= \mathcal{P}_i + \mathcal{R}_{i+1} + \mathcal{Q}^T_{i+1} \\
    \mathcal{P}_1 &= \qdef + \mathcal{Q}^T_1 + \mathcal{Q}_1
\end{align*}

We use $\mathcal{Q}_1$ to denote either the dense caption or image that the agent must classify. We take $\mathcal{R}_4$ as the final response from the model to extract the classification. The text queries for each step are given as follows:

\begin{tcolorbox}
\begin{enumerate}
    \item[$\mathcal{Q}^T_1$:] In accordance with the definition for ``Hazard," Identify any real physically present element that could cause immediate harm. Also consider if people or property are directly in danger.

    \item[$\mathcal{Q}^T_2$:] Independently determine whether any element violates normal expectations for this environment using the definition for ``Anomaly".
    \item[$\mathcal{Q}^T_3$] $=\qtf$
\end{enumerate}
\end{tcolorbox}

\pseudosection{Model Selection} Since our task is designed to evaluate VLM effectiveness for autonomous annotation of hazards, we argue that it is important for models to be both accurate and fast, especially when deployed in real-time applications. Given this, we seek to compare the potential trade-off in accuracy and gain in speed when using smaller, more optimized models vs. larger ones like GPT 4.1 or Mistral Large.  For our study, we use GPT 4.1; Mistral Ministral 3B, 14B, and Large; and Gemini 2.5 Flash and 3 Flash. The primary motivations for selecting these models were to evaluate how models of different sizes perform on this task, as well as to compare how older models perform relative to newer models.

\pseudosection{Baseline Comparison} The most closely related prior work on evaluating whether VLMs can differentiate between safe and hazardous~\cite{bettersafethansorry} considers only binary classification and can be treated as a relevant baseline for our more complex classification. We hypothesize that explicitly separating anomalous from hazardous cases reduces misclassifications between these categories, thereby mitigating the overreaction problem observed in VLM systems~\cite{bettersafethansorry}.
To enable comparison, we adopt their dataset and adapt its annotations to our four-class scheme. Because the original dataset does not include an anomalous category, we map exclusively ``anomalous" cases to ``safe" and ``anomalous–hazardous" cases to ``hazardous" for binary evaluation. However, we keep our prompts the same to still allow models to be more granular in their assignment. This mapping allows direct comparison with the results reported in the baseline study~\cite{bettersafethansorry}.





\section{Results and Analysis}

\begin{table}
\vspace{.5cm}
\setlength{\fboxsep}{3pt}
\begin{tabular}{l *{6}{@{\hskip -3.5pt}S}}
\toprule
\textbf{Model}  & {$\bf P_A$} & {$\bf R_A$} & {$\bf F1_A$} & {$\bf P_H$} & {$\bf R_H$} & {$\bf F1_H$} \\
\midrule
\rowcolor{light_gray}
 \multicolumn{7}{l}{\colorbox{light_gray}{\bf Image Input}} \\
 
 \multicolumn{7}{l}{\emph{\colorbox{white}{Chain of Thought Prompting}}} \\
 
 Ministral 3b & 0.566667 & 0.850000 & 0.680000 & 0.612500 & 0.980000 & 0.753846 \\
 Ministral 14b & 0.510526 & \textbf{0.97} & 0.668966 & 0.603659 & \textbf{0.99} & 0.750000 \\
 Mistral large & 0.612403 & 0.790000 & 0.689956 & 0.664384 & 0.970000 & 0.788618 \\
 Gemini 3 Flash & 0.589928 & 0.820000 & 0.686192 & \textbf{0.73} & 0.960000 & \textbf{0.83} \\
 Gemini 2.5 Flash & 0.605442 & 0.890000 & \textbf{0.72} & 0.680556 & 0.980000 & 0.803279 \\
 Gpt 4.1 & \textbf{0.65} & 0.810000 & \textbf{0.72} & 0.713235 & 0.970000 & 0.822034 \\
  
  \multicolumn{7}{l}{\emph{\colorbox{white}{Few Shot Prompting}}} \\

 Ministral 3b & 0.572650 & 0.690722 & 0.626168 & 0.754902 & 0.793814 & 0.773869 \\
 Ministral 14b & 0.624000 & 0.780000 & 0.693333 & 0.754545 & 0.830000 & 0.790476 \\
 Mistral large & \textbf{0.69} & 0.700000 & 0.696517 & 0.784314 & 0.800000 & 0.792079 \\
 Gemini 3 Flash & 0.658333 & 0.790000 &\textbf{ 0.72} & \textbf{0.82} & 0.880000 & \textbf{0.85} \\
 Gemini 2.5 Flash & 0.604167 & \textbf{0.88} & \textbf{0.72} & \textbf{0.82} & 0.880000 & \textbf{0.85} \\
 Gpt 4.1 & 0.650000 & 0.780000 & 0.709091 & 0.803571 &\textbf{ 0.90} & \textbf{0.85} \\
  
  \multicolumn{7}{l}{\emph{\colorbox{white}{Zero Shot Prompting}}} \\

 Ministral 3b & 0.561404 & 0.390244 & 0.460432 & 0.596491 & 0.944444 & 0.731183 \\
 Ministral 14b & 0.566667 & \textbf{0.86} & 0.682731 & 0.681481 & 0.929293 & 0.786325 \\
 Mistral large & 0.664000 & 0.830000 & \textbf{0.74} & 0.618421 & 0.940000 & 0.746032 \\
 Gemini 3 Flash & \textbf{0.70} & 0.700000 & 0.700000 & \textbf{0.77} & 0.890000 & \textbf{0.82} \\
 Gemini 2.5 Flash & 0.592000 & 0.840909 & 0.694836 & 0.675000 & \textbf{0.95} & 0.790244 \\
 Gpt 4.1 & 0.679612 & 0.700000 & 0.689655 & 0.693431 & \textbf{0.95} & 0.801688 \\
 
\rowcolor{light_gray}
 \multicolumn{7}{l}{\colorbox{light_gray}{\bf Dense Caption Input}} \\
 
  \multicolumn{7}{l}{\emph{\colorbox{white}{Chain of Thought Prompting}}} \\
 Ministral 3b & 0.536765 & 0.730000 & 0.618644 & 0.574850 & 0.960000 & 0.719101 \\
 Ministral 14b & 0.508287 &\textbf{ 0.92} & 0.654804 & 0.611465 & 0.960000 & 0.747082 \\
 Mistral Large & 0.536232 & 0.740000 & 0.621849 & 0.683099 & \textbf{0.97} & 0.801653 \\
 Gemini 3 Flash & 0.612069 & 0.710000 & 0.657407 & \textbf{0.76} & 0.890000 & \textbf{0.82} \\
 Gpt 4.1 & \textbf{0.64} & 0.72 & \textbf{0.68} & 0.70 & 0.96 & 0.81 \\
 
 \multicolumn{7}{l}{\emph{\colorbox{white}{Few Shot Prompting}}} \\

 Ministral 3b & 0.500000 & 0.717391 & 0.589286 & 0.787234 & 0.778947 & 0.783069 \\
 Ministral 14b & 0.626087 & 0.720000 & 0.669767 & \textbf{0.86} & 0.670000 & 0.752809 \\
 Mistral Large & \textbf{0.67} & 0.65 & 0.66 & \textbf{0.86} & 0.76 & 0.81 \\
 Gemini 3 Flash & 0.663551 & 0.710000 & \textbf{0.69} & 0.847826 & 0.780000 & 0.812500 \\
 Gemini 2.5 Flash & 0.589552 & \textbf{0.79} & 0.675214 & 0.826923 & \textbf{0.86} & \textbf{0.84} \\
 Gpt 4.1 & 0.663158 & 0.630000 & 0.646154 & 0.803922 & 0.820000 & 0.811881 \\
 
 \multicolumn{7}{l}{\emph{\colorbox{white}{Zero Shot Prompting}}} \\

 Ministral 3b & 0.562500 & 0.214286 & 0.310345 & 0.666667 & 0.933333 & 0.777778 \\
 Ministral 14b & 0.555556 & 0.750000 & 0.638298 & 0.615385 & 0.888889 & 0.727273 \\
 Mistral Large & 0.603604 & 0.670000 & 0.635071 & 0.671429 & \textbf{0.94} & 0.783333 \\
 Gemini 3 Flash & \textbf{0.67} & 0.520000 & 0.584270 & 0.745455 & 0.820000 & 0.780952 \\
 Gemini 2.5 Flash & 0.650000 & \textbf{0.78} & \textbf{0.71} & \textbf{0.76} & 0.909091 & \textbf{0.83} \\
 Gpt 4.1 & 0.634146 & 0.520000 & 0.571429 & 0.745902 & 0.910000 & 0.819820 \\

\bottomrule
\end{tabular}
\caption{Shows the per-class precision, recall, and f1 scores for anomaly and hazard classes, denoted $P$, $R$, and $F1$ respectively with the subscript indicating the class.}
\label{tab:our-dataset-summary}
\end{table}

\begin{table}[]
\vspace{3.5ex}
\resizebox{0.92\columnwidth}{!}{%
\begin{tabular}{cccccccc}
\multicolumn{1}{c|}{\textbf{Reasoning}} &
  \multicolumn{1}{c|}{\textbf{Inputs}} &
  \multicolumn{6}{c}{\textbf{Hamming Loss}} \\ \hline
\multicolumn{1}{c|}{\multirow{2}{*}{\begin{tabular}[c]{@{}c@{}}Chain-of-\\Thought\end{tabular}}} &
  \multicolumn{1}{c|}{\cellcolor{light_gray}Image} &
  \multicolumn{1}{c|}{\cellcolor{light_gray}0.36} &
  \multicolumn{1}{c|}{\cellcolor{light_gray}0.41} &
  \multicolumn{1}{c|}{\cellcolor{light_gray}0.31} &
  \multicolumn{1}{c|}{\cellcolor{light_gray}0.28} &
  \multicolumn{1}{c|}{\cellcolor{light_gray}--} &
  \cellcolor[HTML]{DDDDDD}\textbf{0.26} \\ 
\multicolumn{1}{c|}{} &
  \multicolumn{1}{c|}{Captions} &
  \multicolumn{1}{c|}{0.41} &
  \multicolumn{1}{c|}{0.41} &
  \multicolumn{1}{c|}{0.34} &
  \multicolumn{1}{c|}{\bf 0.28} &
  \multicolumn{1}{c|}{--} &
  0.29 \\ \hline 
\multicolumn{1}{c|}{\multirow{2}{*}{\begin{tabular}[c]{@{}c@{}}Few\\ Shot\end{tabular}}} &
  \multicolumn{1}{c|}{\cellcolor{light_gray}Image} &
  \multicolumn{1}{c|}{\cellcolor{light_gray}0.32} &
  \multicolumn{1}{c|}{\cellcolor{light_gray}0.28} &
  \multicolumn{1}{c|}{\cellcolor{light_gray}0.26} &
  \multicolumn{1}{c|}{\cellcolor{light_gray}\bf 0.23} &
  \multicolumn{1}{c|}{\cellcolor{light_gray}0.25} &
  \cellcolor[HTML]{DDDDDD}0.24 \\ 
\multicolumn{1}{c|}{} &
  \multicolumn{1}{c|}{Captions} &
  \multicolumn{1}{c|}{0.35} &
  \multicolumn{1}{c|}{0.29} &
  \multicolumn{1}{c|}{0.26} &
  \multicolumn{1}{c|}{\bf 0.25} &
  \multicolumn{1}{c|}{0.27} &
  0.27 \\ \hline 
\multicolumn{1}{c|}{\multirow{2}{*}{\begin{tabular}[c]{@{}c@{}}Zero\\ Shot\end{tabular}}} &
  \multicolumn{1}{c|}{\cellcolor{light_gray}Image} &
  \multicolumn{1}{c|}{\cellcolor{light_gray}0.40} &
  \multicolumn{1}{c|}{\cellcolor{light_gray}0.33} &
  \multicolumn{1}{c|}{\cellcolor{light_gray}0.31} &
  \multicolumn{1}{c|}{\cellcolor{light_gray}\bf 0.24} &
  \multicolumn{1}{c|}{\cellcolor{light_gray}0.31} &
  \cellcolor[HTML]{DDDDDD}0.28 \\ 
\multicolumn{1}{c|}{} &
  \multicolumn{1}{c|}{Captions} &
  \multicolumn{1}{c|}{0.33} &
  \multicolumn{1}{c|}{0.38} &
  \multicolumn{1}{c|}{0.32} &
  \multicolumn{1}{c|}{0.30} &
  \multicolumn{1}{c|}{\bf 0.26} &
  0.29 \\ \hline
 &
  \multicolumn{1}{c}{} &
  \begin{rotate}{325}Ministral 3B\end{rotate} &
  \begin{rotate}{325}Ministral 14B\end{rotate} &
  \begin{rotate}{325}Mistral Large\end{rotate} &
  \begin{rotate}{325}Gemini 3 Flash\end{rotate} &
  \begin{rotate}{325}Gemini 2.5 Flash\end{rotate} &
  \begin{rotate}{325}GPT 4.1\end{rotate} \\
\multicolumn{1}{l}{} &
  \multicolumn{1}{l}{} &
  \multicolumn{1}{l}{} &
  \multicolumn{1}{l}{} &
  \multicolumn{1}{l}{} &
  \multicolumn{1}{l}{} &
  \multicolumn{1}{l}{} &
  \multicolumn{1}{l}{} \\
\multicolumn{1}{l}{} &
  \multicolumn{1}{l}{} &
  \multicolumn{1}{l}{} &
  \multicolumn{1}{l}{} &
  \multicolumn{1}{l}{} &
  \multicolumn{1}{l}{} &
  \multicolumn{1}{l}{} \\
\multicolumn{1}{l}{} &  \multicolumn{1}{l}{} &
  \multicolumn{1}{l}{} &
  \multicolumn{1}{l}{} &
  \multicolumn{1}{l}{} &
  \multicolumn{1}{l}{} &
  \multicolumn{1}{l}{} &
  \multicolumn{1}{l}{}
\end{tabular}%
}
    \caption{Shows the Hamming loss for all models for \textit{image} and \textit{dense caption} input modalities. We omit the results for Gemini 2.5 Flash and chain of thought prompting because the model did not produce valid responses for a significant number of queries. Lower values of hamming loss indicate a lower number of errors in jointly classifying anomalies and hazards. This metric serves to show how well modern VLMs can differentiate the concepts of anomaly and hazard.}
    \label{tab:hamming-loss}
\end{table}

For our evaluation, we first treat each class individually in order to get the per-class performance, rather than performing evaluations jointly for both anomaly and hazard. This allows assigning partial correctness in model outputs. These performance metrics are summarized in~\cref{tab:our-dataset-summary}.

\begin{figure}[t]
    \vspace{0.3cm}
    \centering
    \includegraphics[width=0.9\linewidth]{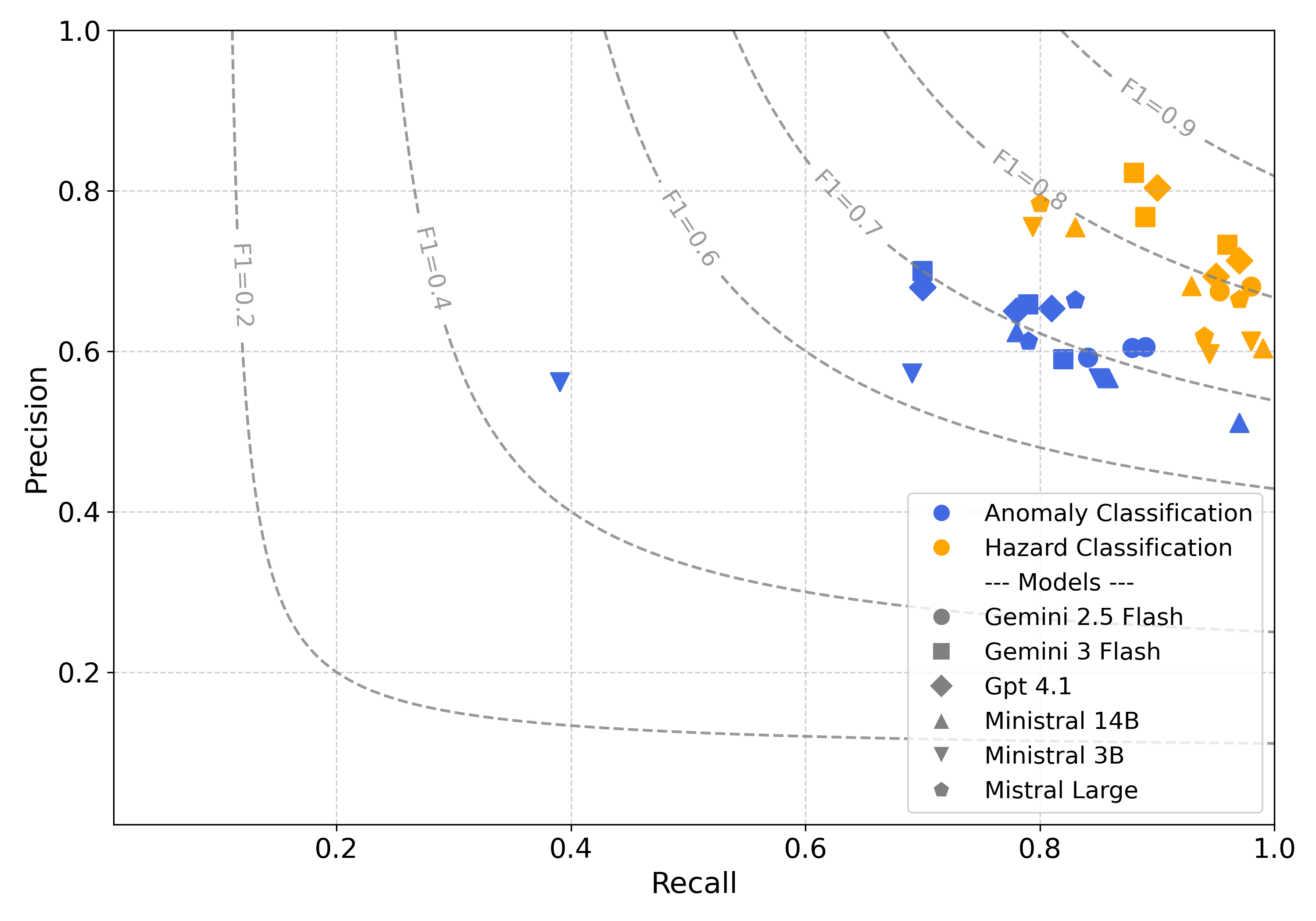}
    \caption{The precision vs. recall graph when using images as the input modality for the agents. Experiment results where an agent did not produce a valid response for more than 10 images per prompting technique were omitted. Generally, the agents perform well when looking at anomaly and hazard classifications independently. The outlier was mistral-3b when using zero-shot prompting.}
    \label{fig:img_prec_v_recall}
\end{figure}

\begin{figure}[t]
    \vspace{0.3cm}
    \centering
    \includegraphics[width=0.9\linewidth]{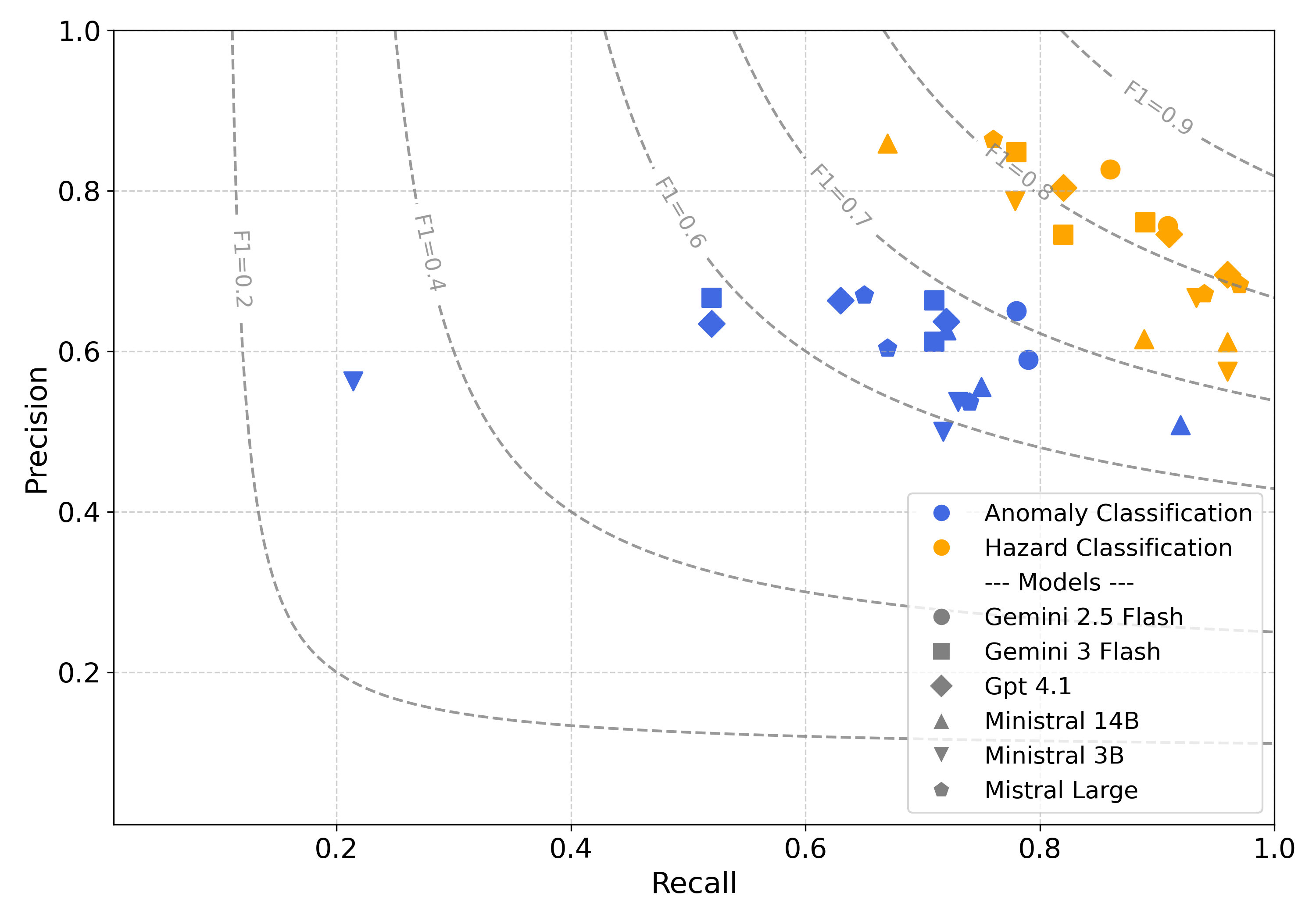}
    
    \caption{The precision vs. recall graph when using dense captions as the input modality for the agents. As with \cref{fig:img_prec_v_recall}, experiments with more than 10 invalid classifications per prompt technique were omitted. Contrary to our expectation, agents performed similarly to when the input modality was an image with some results being worse. A potential cause may be that textual descriptions may be too coarse for the level of information required for the anomaly vs. hazard classification.}
    \label{fig:dense_cap_prec_v_recall}
\end{figure}

\pseudosection{Image vs. Dense Captioning Results} \Cref{fig:img_prec_v_recall,fig:dense_cap_prec_v_recall} show that most models evaluated perform fairly well, especially with hazard classification given F1 scores tended to be above 0.7. As expected, the smallest model(Mistral-3B) is the worst-performing with its F1 score being as low as 0.31 when using the zero-shot prompting method and dense captions as the input modality. Contrary to our expectations, we note that agents tended to perform worse when using dense captions as the input modality with F1 scores skewing lower. We suspect that this is for a few reasons.

It is possible that although models may have the capability to perform the high level reasoning necessary for this classification task, the dense captions may simply not focus enough on the context necessary for the classification. Although the captions were ensured to be accurate for each image, they were made to be neutral about whether the image was hazardous or anomalous. The agent generating the dense captions, Gemini 3 Flash, was given no information about the classification task in order for it to not imbue a bias on other models when they classify anomalies or hazards. As such, these captions may lack the necessary focus to allow all models the best chance to produce accurate results.

Regardless, model performance using images as the input modality indicates that current VLMs are indeed approaching the required level of reasoning capability to effectively classify anomalies and hazards. We observed that even small models like Ministral 3B can, with the right prompting technique, effectively classify anomalies and hazards. For instance, the model achieved an F1 score of 0.68 using chain-of-thought prompting for the anomaly classification and 0.77 in the hazard classification few shot prompting. As shown in \cref{fig:img_prec_v_recall}, agents like Gemini 2.5 and 3 flash reached F1 scores as high as 0.85 on the hazard classification.

It should also be noted that although the few-shot prompt was generally best for classifying hazards, there was no universally best technique across the models for anomaly classifications. Usually, either the chain-of-thought or few-shot prompt gave the best results for the latter classification task depending on the model. For instance, Ministral 3B performed best when using chain-of-thought prompting vs. Ministral 14B that performed better with few-shot prompting. These results suggest that the specific model and prompt can significantly affect performance in deployments like our prior work~\cite{eskandari2025llmsupported}, and factors like the frequency of anomalies or hazards, and the criticality of each class need to be considered when choosing model/prompt combination.

Another interesting feature of both~\cref{fig:img_prec_v_recall} and \cref{fig:dense_cap_prec_v_recall} is how anomaly classification performance is consistently lower than hazard classification performance. This might suggest that semantically, the concept of anomaly is less well-defined. We note that even when annotating our dataset, we tended to have less consensus on the classification for anomalies. These images were carefully considered and excluded where consensus could not be reached.

\pseudosection{Joint Classification Performance} We now consider model performance in jointly classifying images and dense captions into the four categories of our dataset. For this analysis, we use hamming loss~\cite{hamming_loss2014} which averages the number of errors in labeling an image safe, anomalous, hazardous, or both over all input images. These scores are shown in \cref{tab:hamming-loss}, with the lower scores indicating the best performance.

We note that models from the Gemini family and GPT 4.1 consistently perform the best, indicating that these models have the greatest ability to differentiate the concept of an anomaly from a hazard. However, models from the Mistral family had lower performance with worse scores corresponding to smaller models. This implies that when deploying models to classify anomalies and hazards in actual applications, larger models may be more ideal. However, such deployments often require real-time performance or are resource constrained, especially when using edge-computing on robots, underscoring the need for more efficient algorithms

Furthermore, the minimum proportion of errors 0.23 when images and few shot prompting with Gemini 3. This indicates that even though performance can be improved by optimizing the prompt, input, and model selection, performance with current algorithms still needs to be improved when the goal is joint classification and differentiating truly hazardous images from merely anomalous ones. This is especially true for safety-critical deployments like in our previous study~\cite{eskandari2025llmsupported}.
\begin{figure*}[t]
\vspace{0.3cm}
\begin{subfigure}[t]{0.48\linewidth}
    \vskip0pt
    \centering
    \includegraphics[width=\linewidth]{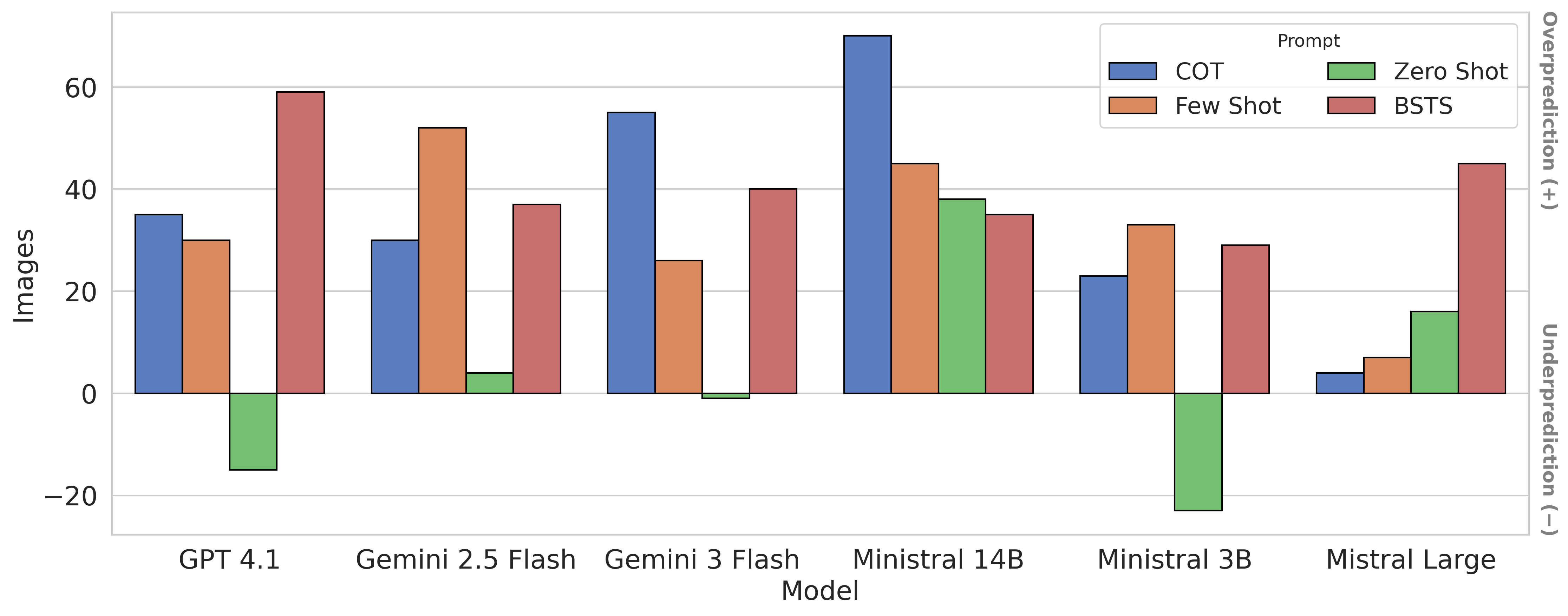}
    \caption{Error in ``Dangerous'' classifications}
    \label{fig:baseline-danger}
\end{subfigure}
\hfill
\begin{subfigure}[t]{0.48\linewidth}
    \vskip0pt
    \centering
    \includegraphics[width=\linewidth]{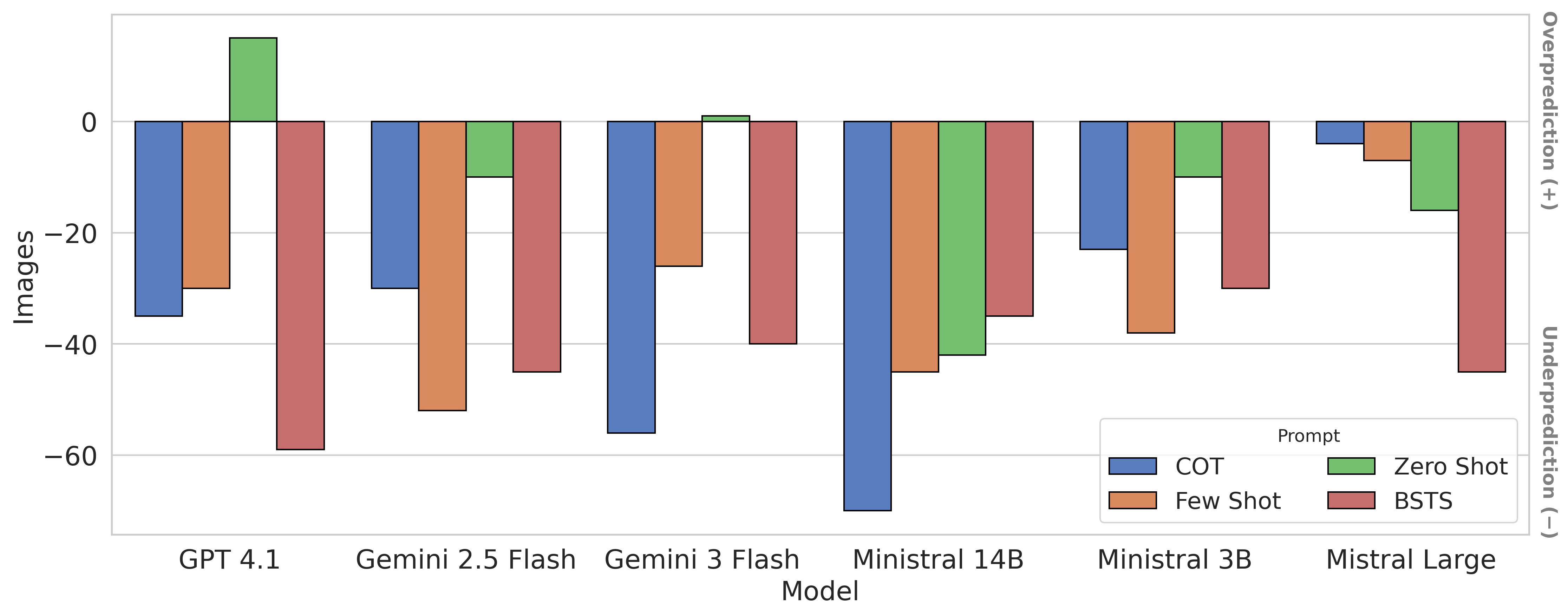}
    \caption{Error in ``Safe'' classifications}
    \label{fig:baseline-safe}
\end{subfigure}
 \caption{Shows how much models underpredicted or overpredicted images classified as ``safe" or ``dangerous'' extracted from the VERI dataset~\cite{bettersafethansorry}. BSTS represents the binary classification prompt used in~\cite{bettersafethansorry}. The dataset had 100 images labeled ``safe.'' Similar to previous findings, most of the prompting techniques tend to under-predict safe images~\cite{bettersafethansorry}, however, our zero-shot prompt is significantly more accurate on average for most models.}
\end{figure*}

\pseudosection{Baseline Comparison} As noted, we observed the best results with images at the input modality. We now review the results when comparing our methodology with previous work on a similar task~\cite{bettersafethansorry}. \Cref{fig:baseline-danger} and \cref{fig:baseline-safe} show the number of over or under-predicted images that each model classified for the ``dangerous'' and ``safe'' categories using the VERI dataset~\cite{bettersafethansorry}. This dataset contains 100 images of each class, all of which were used in raw form rather than as captions.

Unlike in the classification for our dataset, the few-shot technique performs poorly when results are interpreted in a binary manner. This is especially apparent for Ministral 14B. A common reason for misclassifications for many of the over-predicted images seemed to be misinterpretation of the context in images. For instance, one image contained a person with ketchup on them which was interpreted as blood, leading to an incorrect classification. Such errors undermine the system's ability to increase awareness to users when deployed in safety-critical systems; however, false positives such as these are still preferable to false negatives which are more likely to impact safety.

Despite some prompt and model combinations performing poorly, we note that our Zero-Shot prompt proved significantly more accurate than the baseline BSTS prompt~\cite{bettersafethansorry}. A t-test for independence was performed, assuming the null hypothesis that both the Zero-Shot and BSTS prompt accuracies were equal for all models tested. This resulted in a p-value of 0.0022, well past a typical threshold of 0.05 to reject the null hypothesis. Therefore, with the right combination of the model and prompt, the system's performance can be significantly improved over prior work.

\section{Limitations}
Our study identifies systematic deficiencies in joint classification of anomalous and hazardous scenes, but several limitations restrict the generality of the results.

\pseudosection{Dataset size} The dataset contains 610 images, with Hazardous being the smallest category; balanced analysis is limited to roughly 100 samples per class, which amplifies the effect of noise and restricts the diversity of visual patterns the model encounters. Conclusions about hazard recognition should therefore be treated as preliminary, and it is difficult to tell whether observed performance patterns reflect systematic reasoning biases or dataset-specific artifacts. A larger-scale evaluation would yield higher statistical power and enable analysis of generalization.

\pseudosection{Annotations} Ground-truth labels come from a few non-expert annotators, introducing subjectivity for distinctions as subtle as anomalous vs.\ hazardous vs.\ anomalous-hazardous. Our definitions are grounded in OSHA standards~\cite{osha2018hazard}, but personal notions of ``weird,'' ``unsafe,'' and ``dangerous'' vary across individuals. Future work should formalize the annotation protocol via large-scale crowd-sourcing (e.g., Amazon Mechanical Turk) to compute consensus measures and produce probabilistic ground truth, reducing annotator bias.



\section{Conclusion}


This study sought to improve VLM performance in applications such as our prior work~\cite{eskandari2025llmsupported} by creating a dataset of images from safety-critical and benign scenes. We developed a four class classification scheme for images: ``safe,'' ``anomalous,'' ``hazardous,'' and ''anomalous-hazardous.'' Furthermore, we hypothesized that forcing VLMs to separate scenes that are merely anomalous to scenes that are truly hazardous and require immediate action would help focus users' attention to critical details.

Recognizing that VLM performance varies significantly based on model size, prompt, and input, we constructed several sub-tasks for the classification task that used dense captions and raw images as input modalities, and three different prompting techniques for each input modality. When considering per-class classification performance, we found that models perform moderately well with the best models having f1 scores around 0.70 for classifying anomalies and scores as high as 0.85 for classifying hazards. When considering joint classification performance with hamming loss, we found that the best score was 0.23 when using few shot prompting with Gemini 3 and images as the input. This indicated that performance can be improved by tuning pipeline parameters like the model, prompt, and input modality; however, for safety critical tasks like disaster response, current models struggle with the complex reasoning required for the four-class taxonomy.

We found that Choi et al. pursued a similar approach to ours~\cite{bettersafethansorry}, and compared our pipeline with their approach. For this, we collapsed the four-class taxonomy to a binary approach by interpreting anomalies as ``safe,'' and anomalous-hazardous scenes as ``dangerous'' to match their dataset. We found that significant improvements to classification accuracy can be made by tuning the prompt and found that our zero shot prompt was significantly more accurate overall compared to the prior work~\cite{bettersafethansorry}. This may indicate that even if classes are interpreted in a binary manner, forcing models to use the four-class taxonomy helps their reasoning.

Future work should extend this analysis across a wider set of models and incorporate crowd-sourced annotations to build more reliable ground truth. Additional techniques like knowledge graphs of scenes may be incorporated to further assist in reasoning and classification. Such techniques may capture more context than the dense captions that we use for this study and assist models in focusing on only the relevant information for classification. Ultimately, the goal is to guide the development of systems that can distinguish unusual from dangerous conditions with greater consistency and reliability, enabling safer deployments of AI in high-risk environments.



\bibliographystyle{IEEEtranS}
\bibliography{new_references}


\end{document}